# Heart Rate Variability Analysis Using Threshold of Wavelet Package Coefficients

G. Kheder, A. Kachouri, M. Ben Massoued, M. Samet
Laboratory of Electronics and Technologies of Information,
National School of Engineers of Sfax
B.P.W, 3038 Sfax, Tunisia,
gley.kheder@gmail.com

*Abstract*—In this paper, a new efficient feature extraction method based on the adaptive threshold of wavelet package coefficients is presented. This paper especially deals with the assessment of autonomic nervous system using the background variation of the signal Heart Rate Variability HRV extracted from the wavelet package coefficients. The application of a wavelet package transform allows us to obtain a time-frequency representation of the signal, which provides better insight in the frequency distribution of the signal with time. A 6 level decomposition of HRV was achieved with db4 as mother wavelet, and the above two bands LF and HF were combined in 12 specialized frequencies sub-bands obtained in wavelet package transform. Features extracted from these coefficients can efficiently represent the characteristics of the original signal. ANOVA statistical test is used for the evaluation of proposed algorithm.

I. INTRODUCTION

Bioinstrumentation applies the fundamentals of measurement science to biomedical instrumentation. In order to diagnose some diseases, we need to study about the constitution or performance of source from the extracted information. Heart Rate Variability HRV is an immense subject with wide research being conducted all the time around the world. The measurement of the HRV's non-stability presents a challenge to the signal processing techniques, especially in the dynamic conditions of functional testing [1]. The most common mathematical method used to analyze HRV is the Fourier transform, which is limited to stationary signal. The best transformation of the signal expansion is to localize a given basis functions in time and in frequency. The limits of Fourier Transform, while analyzing the functions used are infinitely sharp in their frequency localization. They exist at one exact frequency but have no time localization due to their infinite extends [2]. In fact, to overcome this very limitation, we applied the Wavelet Transform (WT). This transformation is the most efficient method to quantify HRV in non-stationary conditions [1]–[3]–[4]–[5]. In this very research, we are looking for an effective way to analyze the HRV with advanced technique of signal processing. The wavelet transform can be applied to extract the wavelet coefficient of discrete time signals. Compared with the habitual Fourier analysis, wavelet transform shows an idea to watch signals with different scales and to analyze signals with multi-resolution. In our precedent studies we had to use DWT to analyze the HRV, but we encountered frequency decomposition problem [6]. In this paper we use the wavelet package transform (WPT) to decompose the HRV signal into HF and LF frequency ranges. WPT is a more appropriate method to utilize wavelet transform due to the equivalent resolution of gained frequency band.

II. MATERIALS AND METHODS

*A. HRV analysis*

The RR interval variations presented during resting conditions represents a fine tuning of beat-to-beat control mechanisms. Because it helps to evaluate the equilibrium between the sympathetic and parasympathetic influences on heart rhythm, HRV signals analysis is very important and crucial for the study of the Autonomic Nervous System (ANS). The nervous system's sympathetic branch increases the heart rhythm resulting in shorter beat intervals whereas the parasympathetic branch decelerates the heart rhythm leading to longer beat intervals. The spectral analysis of the HRV has led to the identification of two fairly distinct peaks: high (0.15-0.5 Hz) and low (0.05-0.15 Hz) frequency bands. Fluctuations in the heart rate, occurring at the spectral frequency band of 0.15-0.5 Hz, known as high frequency (HF) band, reflect parasympathetic (vagal) activity, while fluctuations in the spectral band 0.05-0.15 hz, known as low frequency (LF) band are linked to the sympathetic modulation, but includes some parasympathetic influence (sympathetic-vagal influences) [3]. It is now established that the level of physical activity is clearly indicated in the HRV power spectrum.

*B. Proposed algorithm*

Fig. 1; show the proposed algorithm for the feature extraction of HRV signal. Feature extraction is a transformation of a pattern from its original form to a new form suitable for further processing. In this study the proposed algorithm is constructed by two principal steps. The first step in performing the feature extraction process should use wavelet package domain. There are many wavelets that can be needed to analyze the signal and extract the feature vector. In this work the Daubechies "db4" wavelet function is used to analyze the signal by WPT. In fact, we obtained maximum energy localization using db4 and db8 when compared to the other

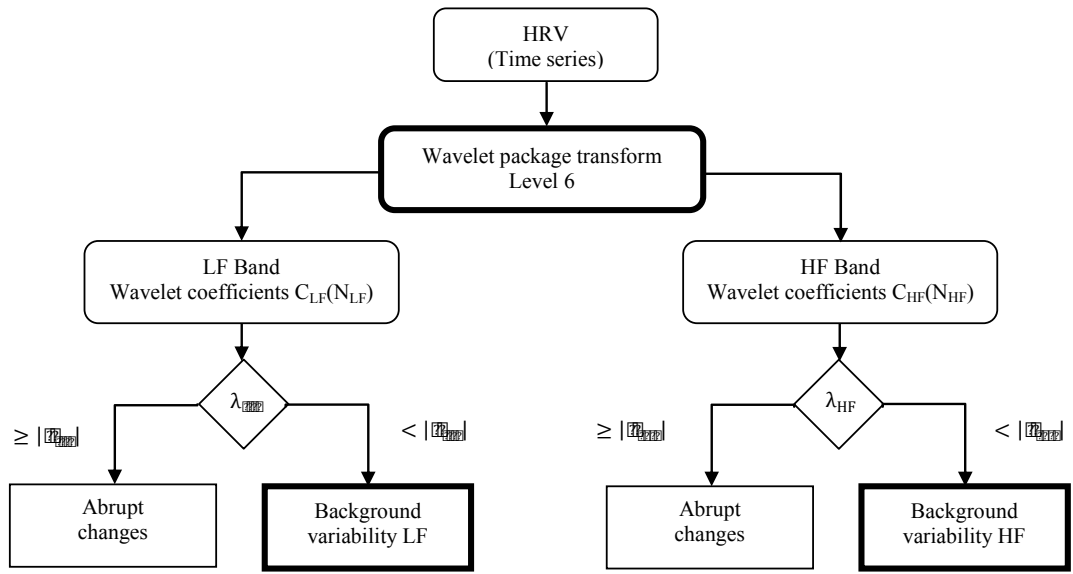

Fig. 1: Proposed algorithm

type of wavelets [8]. The number of decomposition levels is chosen whiten retained based on the dominant frequency components of the signal. The levels are chosen such that those parts of the signal that correlate well with the frequencies required for classification of the signal are retained in the wavelet coefficients.

The second step is: A threshold "λ" was computed according to local characteristics of wavelet coefficients in the two bands LF and HF. Wavelet coefficients were regarded as background variability when they are smaller than the threshold, while they represent truly significant changes when they are greater than the threshold. After thresholding the wavelet coefficients, we get two components to each band, the background variability and other significant signals.

*C. Wavelet theory*

A wavelet family $\psi_{a,b}$ is the set of elemental functions generated by dilatations and translations of a unique admissible mother wavelet $\psi(t)$,

$$\psi_{a,b}(t) = \frac{1}{\sqrt{a}} \psi(\frac{t-b}{a}) \tag{1}$$

The discrete wavelet transform (DWT) achieves this parsimony by restricting the variation in translation and scale, usually to powers of 2. When the scale is changed in powers of 2, the discrete wavelet transform is sometimes termed the dyadic wavelet transform. Discrete wavelet function can be described by (1).

$$\psi_{m,n} = 2^{-m/2} \psi(2^{-m} t - n) \tag{2}$$

Here m is related to a as: $a = 2^m$; b is related to n as $b = n2^m$ and $n, m \in Z$

The wavelet computations are equivalently performed simply using the filtering processes as

$$\phi_{m+1,n}(t) = \frac{1}{\sqrt{2}} \sum_k c_k \phi_{m,2n+k}(k) \tag{3}$$

$$\psi_{m+1,n}(t) = \frac{1}{\sqrt{2}} \sum_k b_k \phi_{m,2n+k}(t) \tag{4}$$

Where $\phi(t)$ is scaling function, $\psi(t)$ is wavelet function, $c_k$ are scaling coefficients, $b_k$ are wavelet coefficients, and k is location index of transform coefficients. Approximation and detail coefficients can be formulized respectively as

$$G_{m+1,n} = \frac{1}{\sqrt{2}} \sum_k c_k A_{m,2n+k} = \frac{1}{\sqrt{2}} \sum_k c_{k-2n} A_{m,k} \tag{5}$$

$$H_{m+1,n} = \frac{1}{\sqrt{2}} \sum_k b_k A_{m,2n+k} = \frac{1}{\sqrt{2}} \sum_k b_{k-2n} A_{m,k} \tag{6}$$

Approximation (A) and detail (D) components is obtained with reconstruction of approximation and detail coefficients as

$$A_M(t) = G_{M,n} \phi_{M,n}(t) \tag{7}$$

$$D_m(t) = \sum_{n=0}^{2^{M-m}-1} H_{m,n} \psi_{m,n}(t) \tag{8}$$

Where M is last decomposition level.

The detail coefficients at all the M levels ($D_1, D_2..., D_M$) and approximate deepest decomposition level (AM). Approximate coefficients often resemble the signal itself.

Initial signal X is reconstruct as

$$X = A_M(t) + \sum_{m=1}^{M} D_m(t), \; m=1, 2, ..., M. \tag{9}$$

Wavelet packet (WP) transform are a generalization of DWT. In WP signal decomposition, both the approximation and detail coefficients are further decomposed at each level. In DWT, detail coefficients are transferred down, unchanged to the next level. However, in WP, all coefficients are decomposed in each stage. WP function includes third additional index as j and is described as (10)

$$W_{m,j,n}(t) = 2^{-m/2} W_j(2^{-m}t - n) \tag{10}$$

Where $j \in N$ denote node index in each m level [9].

### D. Feature extraction and result using WPT

The first stage is the localization of LF and HF bands using wavelet package transform. In literature, critical frequency band that is used determination of sampatho-vagal balance is described as LF and HF that is includes ranges of 0.004-0.15 and 0.15-0.04 Hz, respectively [11]-[12]-[13].

Wavelet packet decomposition at level *j* of HRV signal give $2^j$ sets of sub-band coefficients of length $N/2^j$. The total number of such sets located at the first level to the *j*th level inclusive is $(2^{j+1} - 2)$. The order of these sets at the *j*th level is m= $1,2,\ldots,2^j$. Then, each set of coefficients can be viewed as a node in a binary wavelet packet decomposition tree. Wavelet packet coefficient, $\{P_{j,m}(k) \mid k=1, 2, \ldots, N/2^j\}$, correspond to node (j,m) see TABLE I.

TABLE I. THE FREQUENCY RANGES WITH RESPECT TO NODES

| HRV Bands | Node | Frequency Range (Hz) |
|---|---|---|
|  | (6,0) | 0 – 0,03125 |
|  | (6,1) | 0,03125 – 0,0625 |
| LF | (6,2) | 0,0625 – 0,09375 |
|  | (6,3) | 0,09375 – 0,125 |
|  | (6,4) | 0,125 – 0,15625 |
|  | (6,5) | 0,15625 – 0,1875 |
|  | (6,6) | 0,1875 – 0,21875 |
|  | (6,7) | 0,21875 – 0,25 |
| HF | (6,8) | 0,25 – 0,28125 |
|  | (6,9) | 0,28125 – 0,3125 |
|  | (6,10) | 0,3125 – 0,34375 |
|  | (6,11) | 0,34375 – 0,375 |
|  | (6,12) | 0,375 – 0,40625 |
|  | . | . |
|  | . | . |
|  | . | . |
|  | (6,63) | 1,96875-2 |

These vectors reflect the change of the signal with time in the frequency range of, $\left[\dfrac{(m-1)F_s}{2^{j+1}}, \dfrac{mF_s}{2^{j+1}}\right]$ where $F_s$ is the sampling frequency [9].

The 6 levels decomposition of WP provides high resolution. The obtained frequency bands are too close to LF and HF bands. The resultant resolution of a terminal node is (6,r), r=0, 2, …, 63. The LF band is localized in the nodes (6,1), (6,2), (6,3) et (6,4). However HF band is localized in the nodes (6,5), (6,7), (6,8), (6,9), (6,10), (6,11), (6,12), see Table 1.

### E. Wavelet coefficients threshold

The near-optimally method of wavelet threshold could extract background variability, thanks to Donoho and Johstone [14]-[15]. A threshold "λ" was computed according to local characteristics of wavelet coefficients in the two bands. Wavelet coefficients were regarded as background variability when they are smaller than the threshold, while they represent truly significant changes when they are greater than the threshold. There are two reasons behind this algorithm; the first is, the background activity should widely spread through the whole signal with low amplitudes, the second is, abrupt changes at certain time-points should be localized with high-amplitude wavelet coefficients. After thresholding the wavelet coefficients, we get two components to each band, the background variability and other significant signals.

$$\lambda = h * \sqrt{(2 * \ln(N))} \tag{11}$$

$$h = \frac{MAD}{0.6745} \tag{12}$$

N: is the length of the LF band or HF band
MAD: median absolute deviation

The biggest challenge in feature extraction using wavelet transform is the dominance of random coefficients insignificant. These coefficients may mislead the classification. A new approach is adopted in our work. This approach is based on the idea of extracting a vector describing the background variability of HRV by thresholding wavelet coefficients. Thersholding in general is used in wavelet domain to smooth out or to remove some coefficients of wavelet transform sub-signals of the measured signal.

Great challenge still facing the calculated threshold of wavelet coefficients. Few statistical methods are often used to determine these thresholds. For filtering applications assuming that the analyzed signal contains noise, the universal threshold must still be adjusted.

The entropy in a noisy signal is elevated than in a signal without noise. The correct relationship between the noise and entropy growth is outside scope. To obtain the final threshold value we have multiplied by the so-called scaled median absolute deviation (MAD) computed from the wavelet coefficients of the first level of the transform. The scaled MAD is computed by dividing the MAD by 0.6745. This constant is anticipated in [14] and is based on the wavelet coefficients statistics. In this work MAD is calculated on the various LF and HF bands, for this reason our detection threshold is called local threshold. Moreover the constant mad differs from one segment to another, so it looks like adaptive. In the end our detection threshold "λ" is adaptive local threshold.

III. RESULTS/ DISCUSSION

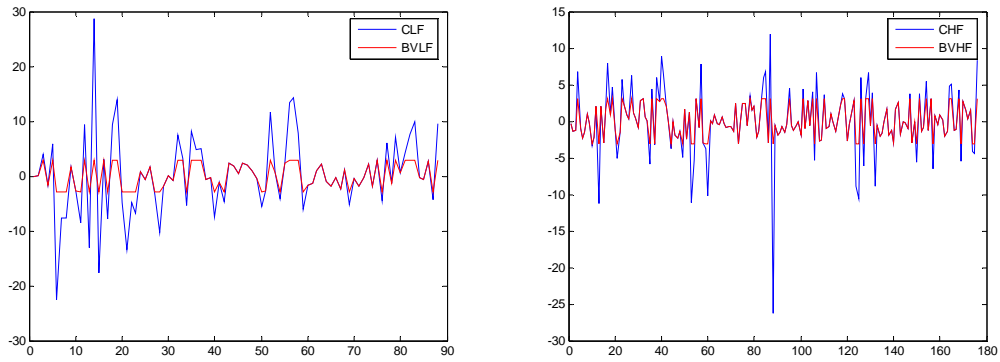

Fig. 2: example of control signal

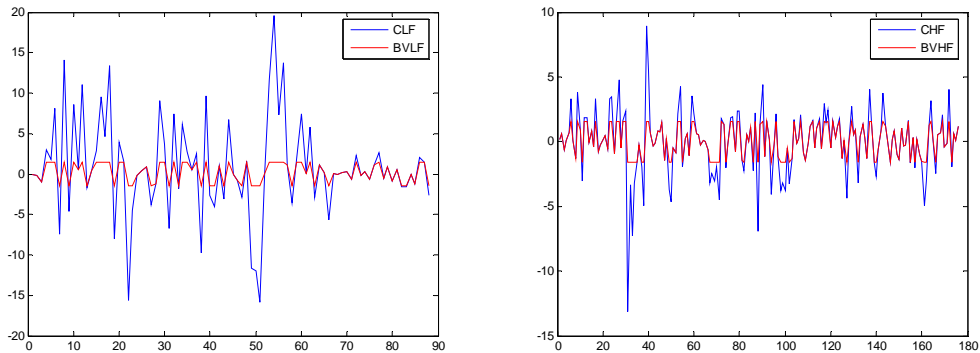

Fig. 3: HRV with VT

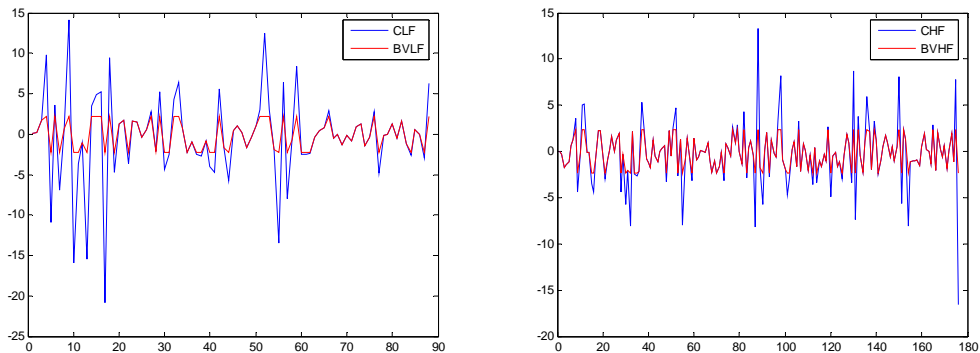

Fig. 4: HRV with VF

The wavelet package coefficients in the finest wavelet domain, manifesting the behavior of parasympathetic and sympathetic components, could be divided into two states: abrupt behavior occurring somewhere in the signal, and background HRV.
The dataset used in this study is obtained from physioBank entitled"Spontaneous Ventricular Tachyarrhythmia Database" [16]. This database contains 135 pairs of RR interval time series, recorded by implanted cardioverter defibrillators in 78 subjects. Each series contains between 986 and 1022 RR intervals. One series of each pair includes a spontaneous episode of ventricular tachycardia (VT) or ventricular fibrillation (VF), and the other is a sample of the intrinsic (usually sinus) rhythm. The ICD maintains a buffer containing the 1024 most recently measured RR intervals. Sampled signals are interpolated using cubic spline interpolation and resample in 4 Hz.

The following three figures illustrate the wavelet coefficients describing the two components LF and HF (CLF and CHF), with the background variability of this components (BVLF and BVHF).

The first empirical interpretation of way of these results presents the distinction of two components after thresholding. Indeed, each branch LF and HF are unscrewed in two vectors built by wavelet coefficients, one illustrates the background variability and the second contain some other elements.

To statistically test whether the different operating conditions produced significant effects on the LF and HF bands of the HRV signals, the method of the Analysis of Variance (ANOVA) was used. ANOVA tests the null hypothesis of equal means between different groups of wavelet coefficients features by analyzing or comparing the sample variance of these groups. Given as recipe, ANOVA involves roughly 5 steps [17]-[18]:

The first shows the source of the variability.

The second shows the Sum of Squares (SS) due to each source.

The third shows the degrees of freedom (df) associated with each source.

The fourth shows the Mean Squares (MS), which is the ratio SS/df.

The fifth shows the critical F ratio, which is the ratio of the mean squares.

Two-way analysis of variance (ANOVA) was used in order to examine statistically significant differences of the background variability between patient with pathologies and control group, and between the different features measurements. The possibility of error in the classification is shown through the computation of the p-value and taken as significant when p<0.05. This method is applied since it is the most common among the clinical community in the statistical assessment of classification studies [19].

TABLE II. RESULTS OF TWO-WAY ANOVA(A,4) TEST, THE MEASUREMENT PARAMETERS ARE; STDLF, MEANLF, STDHF, MEANHF FOR CONTROL, VT AND VF GROUPS.

| Source | SS | df | MS | F | p |
|---|---|---|---|---|---|
| Columns | 215.271 | 3 | 71.7571 | 4.94 | 0.0082 |
| Rows | 35.261 | 2 | 17.6305 | 1.21 | 0.3149 |
| Interaction | 27.162 | 6 | 4.5271 | 0.31 | 0.9247 |
| Error | 348.852 | 24 | 14.5355 | | |
| Total | 626.547 | 35 | | | |

TABLE III. RESULTS OF TWO-WAY ANOVA(A,3) TEST, THE MEASUREMENT PARAMETERS ARE; $E_{LF}$, $E_{HF}$, $R_E$ FOR CONTROL, VT AND VF GROUPS.

| Source | SS | df | MS | F | p |
|---|---|---|---|---|---|
| Columns | 248.64 | 8 | 31.08 | 1.67 | 0.127 |
| Rows | 145.15 | 2 | 72.5734 | 3.9 | 0.0262 |
| Interaction | 974.24 | 16 | 60.8901 | 3.27 | 0.0006 |
| Error | 1004.66 | 54 | 18.6048 | | |
| Total | 2372.69 | 80 | | | |

The energy of wavelet coefficient in the domain of background variability given these parameters:

$E_{LF}$: Energy of wavelet coefficient in LF bands

$E_{HF}$: Energy of wavelet coefficient in HF bands

$R_E$: Ratio of energy $E_{LF}/E_{HF}$;

The sampatho-vagal balance LF/HF ratio and LF and HF energy were significantly different. The interaction effect between the variables is shown in TABLE III, where the sum of squares between the two variables is 974.24, with a mean square of 60.89 (F=3.27, p<0.05). This result shows the interaction effect has significant effect on the wavelet coefficients energy of background variability.

The results of this study showed that wavelet thresholds significantly decreased in patients with VT compared with normal subjects. The ration of the energy composed of wavelet coefficients given a best assessment of VF anomalies in the finest frequency domain of background variability.

VF presents a special condition when wavelet threshold is applied. With chaotic ventricular activity, there is no distinct separation between the irregularly and normal rhythm using wavelet coefficient energy in the background variability (p=0.127) view TABLE III. Finally, we just build a classifier effective process for both pathological VT and VF using wavelet package analysis.

IV. CONCLUSION

Wavelet domain HRV variables provide more specific information about autonomic activity. The WPT method was found to have good time-frequency resolution and give reasonable classification results which compare well with the other approaches.

The results of the ANOVA test showed that the VF anomalies could be well represented by the wavelet coefficients energy. The approaches of wavelet coefficient measurements used in this paper have provided the specification suitable biomarkers that distinguish between signals with pathologies and controls. Thus, our study has established a means of evaluating the features extracted using wavelet coefficient in the domain of background variability in response to HRV diagnosis. The obtained results can be used for diagnosis and classification by comparing with other ailments. It can be speculated that the wavelet threshold, a reliable measurement of HRV with an ability to optimally extract the background component.

## AUTHORS PROFILE

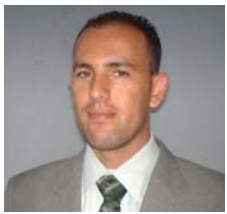

**Gley Kheder** was born in Gafsa, Tunisia in 1978. He received the electrical engineering degree in 2002 and Master degree in electronics and telecommunication in 2003, both from National School of Engineers of Sfax, Tunisia (ENIS). he currently is working toward the Ph.D. degree in electrical ingineering at the same school. Her research interest is to feature extraction of the heart rate variability using advanceds signal processing techniques.

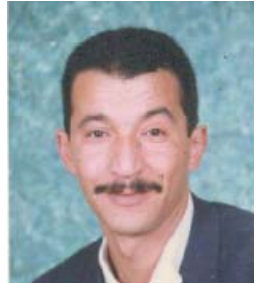

**Abdennaceur Kachouri** was born in Sfax, Tunisia, in 1954. He received the engineering diploma from National school of Engineering of Sfax in 1981, a Master degree in Measurement and Instrumentation from National school of Bordeaux (ENSERB) of France in 1981, a Doctorate in Measurement and Instrumentation from ENSERB, in 1983 and the Habilitation Degree (PostDoctorate degree) in 2008. He " works " on several cooperation with communication research groups in Tunisia and France. Currently, he is Permanent Professor at ENIS School of Engineering and member in the "LETI " Laboratory ENIS Sfax.

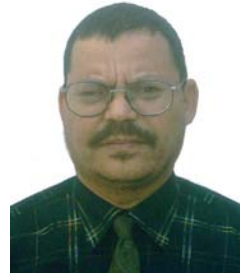

**Mohamed Ben Messaoud** was born in Kebili, Tunisia, on February 2, 1955.
He received the Engineer degree in electric engineering from the University of Sfax, Tunisia, in 1981and the "Docteur Ingenieur" degree in automatic control engineering from the University of Paul Sabatier of Toulouse, France, in 1983.
Since 1983 and the Habilitation Degree (PostDoctorate degree) in 2008, he is with the Department of Electric engineering from the University of Sfax, Tunisia as an Associate Professor. He is also a Member of the Electronic and Information Technology Laboratory for Research on Information theory and Adaptive Control Systems. His current research interests include applied techniques in cardiology, artificial neural network, adaptive observers and their applications

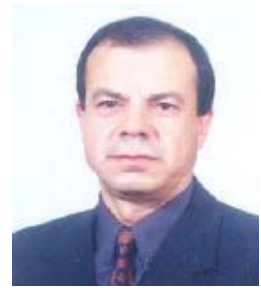

**Mounir Samet** was born in Sfax, Tunisia in 1955. He obtained an Engineering Diploma from National school of Engineering of Sfax in 1981, a Master degree in Measurement and Instrumentation from National school of Bordeaux (ENSERB) of France in 1981, a Doctorate in Measurement and Instrumentation from ENSERB, in 1981 and the Habilitation Degree (PostDoctorate degree) in 1998. He " works " on several cooperation with medical research groups in Tunisia and France. Currently, he is Permanent Professor at ENIS School of Engineering and member in the "LETI" Laboratory ENIS Sfax.